\documentclass[sigconf]{acmart}
\renewcommand\footnotetextcopyrightpermission[1]{}

\AtBeginDocument{%
  }

\setcopyright{acmlicensed}
\copyrightyear{2025}
\acmYear{2025}
\acmDOI{XXXXXXX.XXXXXXX}
\acmConference[MM '25]{Proceedings of the 33rd ACM International Conference on Multimedia}{October 27--31, 2025}{Dublin, Ireland}
\acmISBN{978-1-4503-XXXX-X/2018/06}

\usepackage{multirow}
\usepackage{booktabs}
\usepackage{makecell}
\usepackage{subcaption}
\usepackage{fancybox}
\usepackage{fancyvrb}
\usepackage{bm}
\usepackage{multirow}
\usepackage{booktabs}
\usepackage{bbding}
\usepackage{makecell}
\usepackage{subcaption}
\usepackage{colortbl}
\usepackage{xcolor}
\usepackage{url}
\usepackage{tabularray} 
\usepackage{pifont} 
\usepackage{graphicx} 
\usepackage{makecell} 
\usepackage{wrapfig} 
\usepackage{flushend} 
\definecolor{softred}{RGB}{255, 178, 178}  
\definecolor{softorange}{RGB}{255, 218, 179} 
\definecolor{softyellow}{RGB}{255, 244, 191} 
\usepackage{xcolor}

\begin{document}


\title{A Neural Representation Framework with LLM-Driven Spatial Reasoning for Open-Vocabulary 3D Visual Grounding
}



\author{Zhenyang Liu}
\email{lzyzjhz@163.com}
\affiliation{%
  \institution{Fudan University, Shanghai Innovation Institute}
  \city{Shanghai}
  \country{China}
}

\author{Sixiao Zheng}
\email{sxzheng18@fudan.edu.cn}
\affiliation{%
  \institution{Fudan University, Shanghai Innovation Institute}
  \city{Shanghai}
  \country{China}
}

\author{Siyu Chen}
\email{siyu_chen279@163.com}
\affiliation{%
  \institution{Zhejiang University}
  \city{Hangzhou}
  \state{Zhejiang}
  \country{China}
}

\author{Cairong Zhao}
\email{zhaocairong@tongji.edu.cn}
\affiliation{%
  \institution{Tongji University}
  \city{Shanghai}
  \country{China}
}

\author{Longfei Liang}
\email{longfei.liang@neuhelium.com}
\affiliation{%
  \institution{NeuHelium Co., Ltd}
  \city{Shanghai}
  \country{China}
}

\author{Xiangyang Xue}
\authornote{Corresponding authors.}
\email{xyxue@fudan.edu.cn}
\affiliation{%
  \institution{Fudan University}
  \city{Shanghai}
  \country{China}
}

\author{Yanwei Fu}
\authornotemark[1]
\email{yanweifu@fudan.edu.cn}
\affiliation{%
  \institution{Fudan University, Shanghai Innovation Institute}
  \city{Shanghai}
  \country{China}
}

\renewcommand{\shortauthors}{Trovato et al.}

\begin{abstract}
Open-vocabulary 3D visual grounding aims to localize target objects based on free-form language queries, which is crucial for embodied AI applications such as autonomous navigation, robotics, and augmented reality. Learning 3D language fields through neural representations enables accurate understanding of 3D scenes from limited viewpoints and facilitates the localization of target objects in complex environments.
However, existing language field methods struggle to accurately localize instances using spatial relations in language queries, such as ``the book on the chair.'' This limitation mainly arises from inadequate reasoning about spatial relations in both language queries and 3D scenes. In this work, we propose \textbf{SpatialReasoner}, a novel neural representation-based framework with large language model (LLM)-driven spatial reasoning that constructs a visual properties-enhanced hierarchical feature field for open-vocabulary 3D visual grounding.
To enable spatial reasoning in language queries, SpatialReasoner fine-tunes an LLM to capture spatial relations and explicitly infer instructions for the target, anchor, and spatial relation. To enable spatial reasoning in 3D scenes, SpatialReasoner incorporates visual properties (opacity and color) to construct a hierarchical feature field. This field represents language and instance features using distilled CLIP features and masks extracted via the Segment Anything Model (SAM). The field is then queried using the inferred instructions in a hierarchical manner to localize the target 3D instance based on the spatial relation in the language query. Notably, SpatialReasoner is not limited to a specific 3D neural representation; it serves as a framework adaptable to various representations, such as Neural Radiance Fields (NeRF) or 3D Gaussian Splatting (3DGS). Extensive experiments show that our framework can be seamlessly integrated into different neural representations, outperforming baseline models in 3D visual grounding while empowering their spatial reasoning capability. Project Homepage: \href{https://zhenyangliu.github.io/SpatialReasoner/}{ZhenyangLiu.github.io/SpatialReasoner}.

\end{abstract}




\maketitle

\section{Introduction}
\label{p1}
Open-vocabulary 3D visual grounding~\cite{peng2023openscene, lu2023ovir} aims to localize a target object in a 3D scene using free-form natural language descriptions, enabling humans to interact with the real world through open-ended language. Spatial reasoning is essential for this task, offering new opportunities for human-robot interaction~\cite{sheridan2016human, goodrich2008human}.
\begin{figure}[t]
	\centering
    \includegraphics[width=0.5\textwidth]{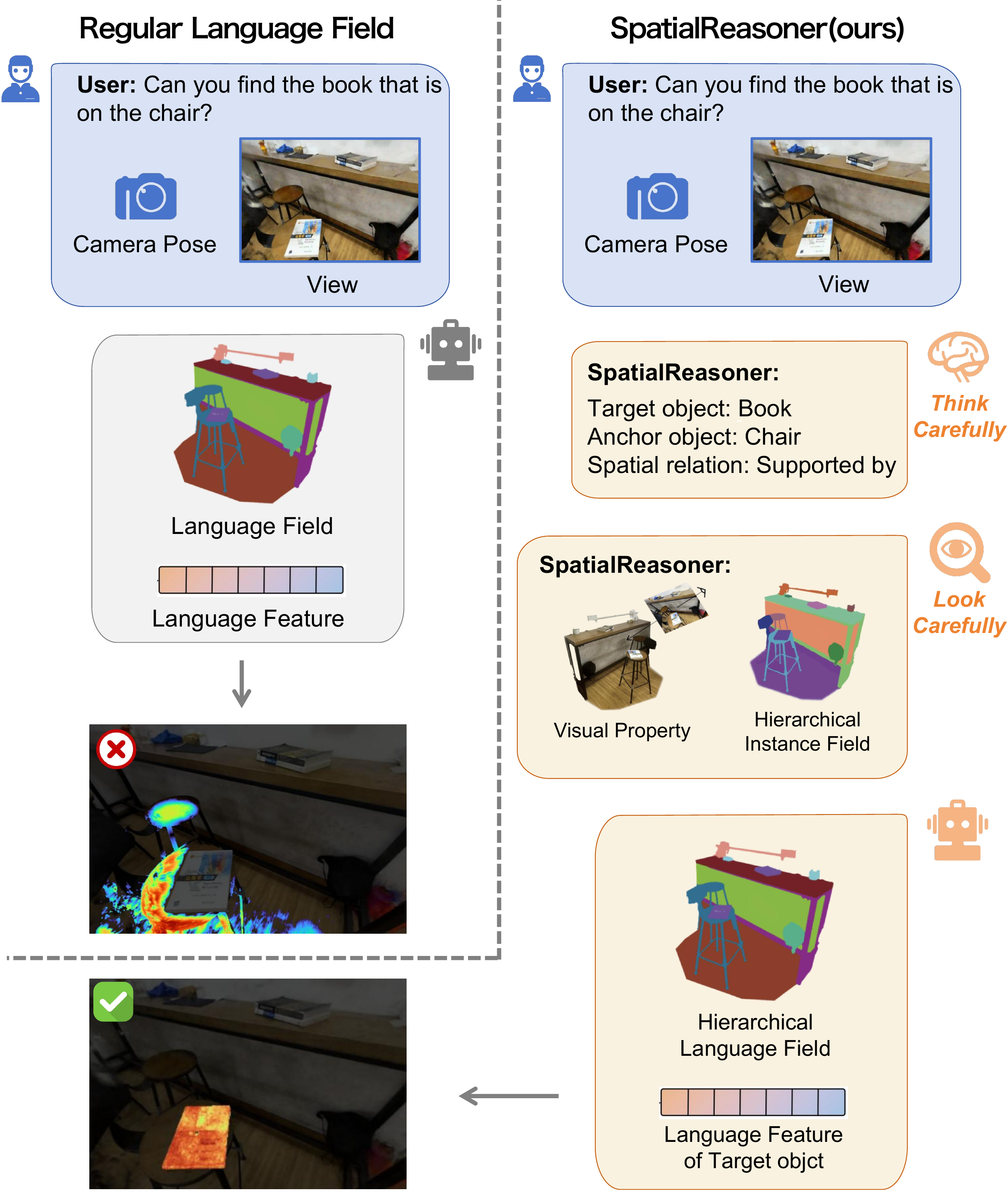}
	\caption{We propose \textit{SpatialReasoner} for neural representation: prior language field methods localize instances directly from complex user queries but fail to capture spatial relations in both the language query and the environment \textit{(left)}. Our \textit{SpatialReasoner} instead utilizes a large language model (LLM) and a hierarchical feature field to think and look ``step by step'' \textit{(right)}. Crucially, reasoning through an LLM—such as spatial relation decomposition—along with hierarchical language and instance fields, allows it to ``think carefully'' and ``look carefully'' before localizing the target instance.}
	\label{1}
    \vspace{-0.17in}
\end{figure}
As shown in Figure~\ref{1}, imagine a robot navigating a room and being asked, ``Can you find the book that is on the chair?'' The robot must interpret the spatial relationship between the book and the chair within the 3D environment, requiring both an understanding of the language and reasoning about spatial relations in real-time. This complex task demands that the robot not only comprehend language but also interprets spatial relations within both the language queries and the environment to precisely identify objects.

Recently, neural representations, such as Neural Radiance Fields (NeRF)\cite{mildenhall2021nerf} and 3D Gaussian Splatting (3DGS)\cite{kerbl20233d}, have emerged as powerful techniques for capturing complex scene structures from posed images. Recent advancements in neural representations have spurred extensive research on open-vocabulary 3D visual grounding with language fields~\cite{kerr2023lerf, zhang2024open, qin2023langsplat}. LERF~\cite{kerr2023lerf} introduces a novel approach to integrating language embeddings, extracted from 2D pretrained vision-language models such as CLIP~\cite{radford2021learning}, into NeRF-based 3D scene representations. LangSplat~\cite{qin2023langsplat} employs 3DGS to build 3D neural representations and incorporates the semantic hierarchy from the Segment Anything Model (SAM)~\cite{kirillov2023segment}.

However, existing language field methods lack spatial reasoning capabilities, making it challenging to accurately localize target instances based on spatial relationships in language queries. As shown in Figure~\ref{1}, regular language field methods localize objects directly based on complex user queries but fail to perceive spatial relationships in both the language query and the environment. Consequently, a localization error occurs: the \textit{chair} is mislocalized, and the \textit{book} specified by the user is not detected. This limitation mainly arises from inadequate reasoning about spatial relations in both language queries and 3D scenes:
\textbf{(1) \textit{Lack of spatial reasoning within the language query.}} Existing language field methods employ the CLIP model for instruction comprehension. However, since the CLIP model is primarily trained on short texts, these methods struggle to accurately capture implicit spatial relations in complex language queries.
\textbf{(2) \textit{Lack of spatial reasoning within the 3D scene.}} Existing language field methods primarily focus on constructing language fields that capture object semantics alone. This limitation impairs spatial reasoning in 3D scenes, making it difficult to differentiate between objects that share the same concept.

In this work, we propose the SpatialReasoner, a novel neural representation-based framework with LLM-driven spatial reasoning that constructs a visual properties-enhanced hierarchical feature field for open-vocabulary 3D visual grounding.
The SpatialReasoner incorporates spatial reasoning into both the 3D scene and language query, enabling accurate localization of specific instances based on the spatial relations described in the query. 
Figure~\ref{1} illustrates the process of spatial reasoning in our SpatialReasoner.
The SpatialReasoner utilizes a large language model (LLM) and a hierarchical feature field to think and look ``step by step''.
Crucially, reasoning through an LLM like spatial relation decomposition in addition to visual properties-enhanced hierarchical feature field to ``think carefully'' and ``look carefully'' before localizing the target instance.

Specifically, we fine-tune an LLM on the Sr3D and Sr3D++~\cite{achlioptas2020referit3d} to enhance its reasoning capabilities, enabling decomposition of spatial relations in language instructions. The fine-tuned model parses complex queries into targets, anchors, and spatial relations. To facilitate 3D spatial reasoning, the SpatialReasoner incorporates visual attributes (opacity and color) to build a hierarchical feature field combining language and instance features.
We first use SAM to extract 2D masks for all objects in the training dataset.
The physical scales of objects are obtained by deprojecting the extracted masks into 3D space using depth information from scene reconstruction.
SpatialReasoner incorporates visual properties derived from scene reconstruction using neural representations.
The position and physical scale, along with the visual properties, are then used to construct a hierarchical feature field that includes both a language field and an instance field.
Both fields are optimized with distilled CLIP features and extracted masks.
The feature field is then queried hierarchically using the inferred instructions, activating potential target and anchor candidates.
SpatialReasoner subsequently constructs an instance graph to further refine these candidates.
Finally, by explicitly verifying whether the activated targets and anchors satisfy the spatial relationships, we precisely localize the specific 3D instance referenced in the language query.
Notably, the proposed SpatialReasoner is not restricted to a specific 3D neural representation; it serves as a framework adaptable to various neural representations, such as NeRF and 3DGS.

To fully evaluate the effectiveness of SpatialReasoner, we conduct extensive experiments on the challenging LERF~\cite{kerr2023lerf}, Replica~\cite{straub2019replica}, and our newly developed Re3D datasets.
To assess the generality of SpatialReasoner, we apply it to various 3D neural representations, including NeRF~\cite{mildenhall2021nerf}, Instant-NGP~\cite{muller2022instant}, and 3DGS~\cite{kerbl20233d}.
We evaluate its performance in 3D visual grounding and spatial reasoning capabilities through comparative experiments on the LERF, Replica, and Re3D datasets, which include complex scenes with diverse objects that share identical semantics but distinct spatial relationships.
Extensive experiments show that SpatialReasoner can be seamlessly integrated into various 3D neural representations. It not only outperforms baseline and previous state-of-the-art methods in 3D visual grounding but also achieves spatial reasoning beyond the capability of language field-based methods, accurately localizing specific instances based on spatial relations in language queries.
The contributions of SpatialReasoner are summarized as follows:
\begin{itemize}
    \item \textbf{(1) \textit{Empowering Language Field with Spatial Reasoning for 
    3D Visual Grounding}}: We overcome the limitation of language field-based open-vocabulary 3D visual grounding, which struggles to localize instances using spatial relations in language queries, by introducing a visual properties-enhanced hierarchical feature field for robust spatial reasoning and accurate grounding.

    \item \textbf{(2) \textit{A Novel SpatialReasoner Framework}}: The proposed SpatialReasoner leverages an LLM for spatial relation decomposition, alongside a visual properties-enhanced hierarchical feature field for spatial reasoning, to ``think carefully'' and ``look carefully'', enabling accurate step-by-step localization of target instances through explicit spatial reasoning.
    \item \textbf{(3) \textit{
    Outstanding Generality and Performance}}: Extensive experiments demonstrate that our method can be seamlessly integrated into diverse 3D neural representations, outperforming baseline models in 3D visual grounding and empowering their spatial reasoning capabilities.	
\end{itemize}

\begin{figure*}
	\centering
	\includegraphics[width=0.95\textwidth]{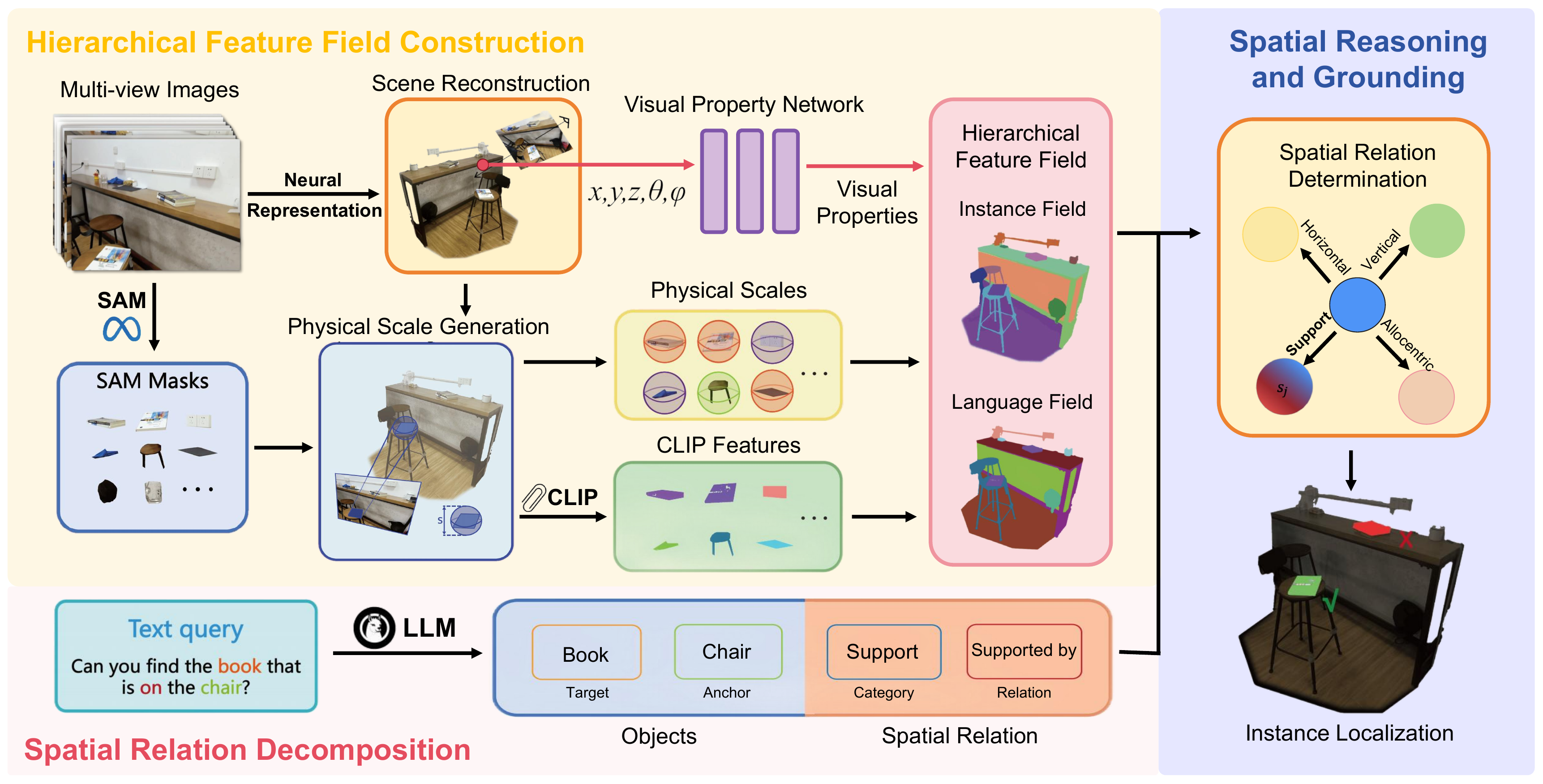}
    \vspace{-0.1in}
	\caption{\textbf{The overall pipeline of SpatialReasoner framework.} SpatialReasoner fine-tunes an LLM to decompose language queries into targets, anchors, and spatial relations. It first employs SAM to generate 2D masks for diverse instances in the training dataset. Using a neural representation model (e.g., NeRF or 3DGS) trained on multi-view images, it obtains instance scales via depth deprojection. By integrating visual properties (opacity and color) from scene reconstruction, SpatialReasoner constructs a hierarchical feature field that combines CLIP-extracted language features and mask-extraceted instance features. Reasoned instructions then query these fields to identify target and anchor candidates. By analyzing spatial relations within the query and the 3D scene, SpatialReasoner precisely localizes the referenced 3D instance.}
    \vspace{-0.10in}
	\label{2}
\end{figure*}

\section{Related Work}
\label{p2}
\noindent\textbf{Visual Grounding on 2D images.} Visual grounding with natural language prompts is essential for robotics and augmented reality, with open-vocabulary localization being a major research focus. Datasets~\cite{zhou2019grounded, krishna2017visual, mao2016generation, yu2016modeling} provide annotated image region descriptions. LSeg~\cite{li2022language} uses a 2D image encoder to generate pixel-wise embeddings that align with CLIP text embeddings. CRIS~\cite{wang2022cris} and CLIPSeg~\cite{luddecke2022image} create relevance maps from query CLIP embeddings. Some methods, like OpenSeg~\cite{ghiasi2022scaling} and ViLD~\cite{gu2021open}, predict masks and use CLIP for open-vocabulary classification.
Unlike feature matching methods for visual grounding, our work focuses on the localization in 3D scenes with language queries.

\noindent\textbf{Visual Grounding on 3D point clouds.} 
Visual grounding on 3D point clouds has attracted growing interest due to its applications in 3D understanding and captioning~\cite{ha2022semantic, hong20233d}, autonomous driving~\cite{novosel2019boosting}, virtual reality~\cite{deng2022fov}, and notably robotic navigation and manipulation~\cite{shah2021ving, lin2023pourit}. Chen et al.~\cite{chen2020scanrefer} introduced the ScanRefer dataset and an end-to-end grounding-by-detection framework. ReferIt3D~\cite{achlioptas2020referit3d} proposed Nr3D and Sr3D datasets, differing by using ground-truth rather than predicted bounding boxes. InstanceRefer~\cite{yuan2021instancerefer} employed a pretrained segmentation model with a handcrafted language parser to select candidate boxes. Nonetheless, these 3D point cloud methods heavily depend on extensive annotations and lack effective integration of 2D foundation model knowledge for zero-shot visual grounding.

\noindent\textbf{3D Language Fields.}
Neural representations~\cite{vora2021nesf,liu2023instance,mildenhall2021nerf,muller2022instant,kerbl20233d} have recently advanced scene reconstruction~\cite{zhang2022nerfusion,remondino2023critical} and novel view synthesis~\cite{tancik2022block,niedermayr2024compressed}. Extending 2D foundation models across views, 3D language fields leverage CLIP features for open-vocabulary 3D visual grounding. Semantic-NeRF~\cite{vora2021nesf} integrates semantics into appearance and structure, while LERF~\cite{kerr2023lerf} projects 2D CLIP embeddings into 3D via NeRF to construct language models within this framework. LangSplat~\cite{qin2023langsplat} pioneered 3D Gaussian Splatting for language fields, and ReasonGrounder~\cite{liu2025reasongrounder} exploits hierarchical 3D Gaussian features for visual grounding and reasoning. Nonetheless, these approaches struggle with precise instance localization due to inadequate spatial relation reasoning in language and 3D representations. In contrast, our SpatialReasoner employs a large language model (LLM) to decompose spatial relations and integrates visually enhanced hierarchical feature fields, enabling stepwise, accurate localization via spatial reasoning.

\section{Method}
\label{p3}
As shown in Figure~\ref{2}, we introduce SpatialReasoner, a neural representation-based framework that leverages an LLM for spatial relation decomposition in language queries, alongside visual properties-enhanced hierarchical feature fields for spatial reasoning and open-vocabulary 3D visual grounding.

\subsection{Preliminary: Language Fields}  
To embed language features into spatial points within neural representations such as NeRF or 3DGS, language fields~\cite{kerr2023lerf,qin2023langsplat} define a mapping function $F$ that takes an input position and volume scale to generate the corresponding language embeddings. 
Given $N$ calibrated input views with known pose information $\{I_i, P_i\}_{i=1}^N$, a 2D vision-language model extracts the corresponding ground-truth 2D features.	
Each expected pixel-wise language embedding $\phi_{\rm lang}$ is obtained by casting a ray $\vec{x}(t) = \vec{o} + t\vec{d}$.	
The language field samples the initial volume scale $s(t)$ for each spatial point along the ray based on focal length and sampling distance from the ray origin.	
Using the same rendering weights as in volume rendering theory~\cite{brebin1998volume}, defined as $T(t) = \int_{t} \exp(-\sigma(s))ds$ and $\omega(t) = \int_{t} T(t) \sigma(t) dt$, the expected pixel-wise language embedding $\phi_{\rm lang}$ is computed as follows:	
	\begin{equation}\label{eq1}
		\begin{aligned}
			&\hat{\phi}_{\rm lang} = \int_{t} \omega(t) F(\vec{x}(t), s(t))dt  \\
			&\phi_{\rm lang} = \hat{\phi}_{\rm lang} / \lVert\hat{\phi}_{\rm lang} \rVert
		\end{aligned}
	\end{equation}
	where $\hat{\phi}_{\rm lang}$ denotes raw language rendering outputs, and $\phi_{\rm lang}$ denotes the final normalized language embedding. This method combines 2D multimodal features to estimate pixel-level 2D representations, optimized by minimizing the loss between $\phi_{\rm lang}$ and the 2D ground truth features.

During the querying stage, the language field computes the CLIP embedding of the language query, $\phi_{\text{quer}}$, and the canonical phrases, $\phi_{\text{canon}}^i$, to determine the appropriate volume scale for language embeddings. The relevance score $R$ is then computed using Eqn.~\ref{eq2}. 
\begin{equation}\label{eq2}
	R = \min_i \frac{\exp({\phi}_{\rm lang} \cdot {\phi}_{\rm quer})}{\exp({\phi}_{\rm lang} \cdot {\phi}_{\rm canon}^i)+ \exp({\phi}_{\rm lang} \cdot {\phi}_{\rm quer})}
\end{equation}
Language fields, commonly used in existing methods~\cite{kerr2023lerf, qin2023langsplat, kim2024garfield}, is adopted by SpatialReasoner to lift 2D features into 3D within neural representations, enabling precise spatial reasoning and grounding.

\subsection{Spatial Relation Decomposition}\label{3.2}
Given a language description, such as ``Can you find the book that is on the chair?'', humans naturally decompose this description
into distinct semantic directives, guiding us first to localize
the ``chair'' (\textit{anchor}),  followed by transferring our focus from the ``chair'' to the ``book'' (\textit{target}), facilitated by the intervening spatial relation ``supported-by'' (\textit{spatial relation}).
Inspired by this cognitive process, we introduce the LLM-driven method to \textbf{``think carefully''}: 
decomposing the given language description into a series of instructions denoted as $\{I_i\}_{i=1}^n$. 

To enable spatial reasoning in language queries, we fine-tune large language models (LLMs) such as ChatGPT\cite{achiam2023gpt} to parse descriptions into key semantic components: the \textit{anchor}, \textit{target}, and their \textit{spatial relation}. These parsed concepts are encoded into embeddings that serve as explicit instructions. This fine-tuning allows SpatialReasoner to accurately interpret and reason about implicit spatial relationships in complex queries. Moreover, the number of instructions (\textit{n}) is adapted per benchmark; for instance, in Sr3D and Sr3D++\cite{achlioptas2020referit3d}, \textit{n} is set to 3 to separately guide reasoning about the \textit{target} category, \textit{anchor} category, and their \textit{spatial relation}. Further details on instruction generation are provided in the Supplementary Material.

\subsection{Hierarchical Feature Field Construction}\label{3.3}
To enable spatial reasoning in 3D scenes and \textbf{``look carefully''}, SpatialReasoner integrates visual properties (opacity and color) to construct a hierarchical feature field that consists of both language and instance feature fields. Given the multi-view images, SpatialReasoner first trains a neural representation model (e.g., NeRF or 3DGS) to extract visual properties, calculate physical scales, and represent the geometric appearance within the hierarchical field.

\noindent \textbf{Visual Property Extraction.}
Existing language field methods encode language embeddings solely via spatial coordinates and volume scales, limiting their capacity to represent complex features such as blurred boundaries, visually similar objects, and transparent surfaces. This constraint impairs embedding quality and instance localization accuracy. Opacity and color at each spatial point provide rich geometric and appearance cues closely linked to language semantics. Opacity reflects material and structural properties, aiding classification, while color captures subtle visual details. By incorporating color into the language embedding, SpatialReasoner enhances discrimination of fine visual features, thereby improving accuracy and contextual understanding in 3D visual grounding.
To enhance the quality of hierarchical features and improve the accuracy of 3D visual grounding, SpatialReasoner incorporates opacity and color as visual properties in the construction of the hierarchical feature field. Based on the trained neural representation model,
the opacity $\sigma$ and color $c$ can be extracted through the visual property network $F_v$, which takes in the 3D coordinate $\vec{x}(t)$ and the viewing direction $(\theta,\phi)$:
\begin{equation}\label{eq3}
	\begin{aligned}
			& (\sigma, c) = F_v(\vec{x}(t), \theta,\phi).
		\end{aligned}
	\end{equation} 
SpatialReasoner relies on the neural representation to determine how to incorporate the visual properties (opacity and color) into the hierarchical feature field, including instance and language features.
The construction of the two fields below illustrates these details.	

\noindent \textbf{Supervision Generation.}
SpatialReasoner first utilizes the automatic mask generator of SAM~\cite{kirillov2023segment} to generate object masks from training views, which are then leveraged to assign physical scales and acquire pixel-aligned features. 
These masks are then filtered based on confidence, and nearly identical ones are deduplicated to generate mask candidates $\{M_1, M_2, ..., M_n\}$. These candidates may overlap or encompass each other. 
Using the trained neural representation model, SpatialReasoner renders the depth of each mask's pixels.	
Given the ray origin, the pixels of each mask can be deprojected onto 3D points using focal length and depth.	
By calculating the \textit{standard deviation} of these points, we can estimate the physical scales $\{S_1, S_2, ..., S_n\}$ for the objects. 
As a foundational multimodal model for text–vision interaction, CLIP~\cite{radford2021learning} uses an image encoder to extract visual features and a text encoder to extract textual features.	
We then extract CLIP features $\{\phi_1, \phi_2, \dots, \phi_n\}$ for each segmented region across multiple training views to ensure multi-view consistency.	
Thus, for each mask candidate, we obtain the associated triplet $\{M_i, S_i, \phi_i\}$.

\noindent \textbf{Language Field.} 3D visual grounding requires the model to localize a target object based on natural language queries.	
SpatialReasoner achieves this by constructing a language field within a hierarchical feature field.	
After obtaining the language embeddings ${\phi_i}$ and 3D physical scales ${S_i}$ from 2D posed images, SpatialReasoner constructs a language field to model the relationships between 3D points and 2D pixels.	
Specifically, a ray $\vec{x}(t) = \vec{o} + t\vec{d}$ is cast from a pixel in the mask candidate $M_k$ with physical scale $S_k$.	
SpatialReasoner incorporates the visual properties into the construction of language field.
SpatialReasoner models a language field $F_l$ that maps the input coordinate {$\vec{x}(t)$}, physical scale $S_k$, and visual properties $\{\sigma, c\}$ to the 3D language embeddings $F_l(\vec{x}(t), S_k, \sigma, c)$.
It is important to note that {$(\vec{x}(t), S_k) \in \mathbb{R} ^ {192}$ and $(\sigma, c) \in \mathbb{R} ^ 4$,} with the dimension of $(\sigma, c)$ being much smaller than that of $(\vec{x}(t), S_k)$. Therefore, introducing density and color will not significantly increase the computational burden.	 
Using the same rendering weights based on density $\sigma$ as the neural representation, $w(t) = \int_{t} \exp(-\sigma(s))ds$, each expected pixel language feature $\hat{\phi}$ is rendered using $F_l(\vec{x}(t), S_k, \sigma, c)$:
\begin{equation}\label{eq4}
	{\hat{\phi}} = \int_{t} \omega(t) F_l(\vec{x}(t), S_k, \sigma, c)dt  
\end{equation}
where $F_l(\vec{x}(t), S_k, \sigma, c)$ denotes the language feature of spatial points on the ray. By optimizing the language loss $L_l = - \lambda_{l} \hat\phi_{k} \cdot \phi_{k}$, where $\lambda_{l}$ is a hyperparameter, the language field learns to project 2D CLIP embeddings into 3D space.	

\noindent \textbf{Instance Field.} The real world encompasses numerous complex scenarios, where instances sharing identical language features must be distinguished. 
For instance, while ``book on the table'' and ``book on the chair'' both involve the language feature ``book'', they represent clearly distinct instances.	
To achieve this, SpatialReasoner constructs a hierarchical instance field to differentiate instances with identical language features and further refine localization precision.	
Similar to the language field, the instance field defines an instance mapping function $F_{in}$, which maps the input spatial coordinate $\vec{x}(t)$, physical scale $S_k$, and visual properties ${\sigma, c}$ to the 3D instance embedding $F_{in}(\vec{x}(t), S_k, \sigma, c)$.	
These 3D instance embeddings can be rendered using the same rendering weights in neural representations, yielding the pixel instance feature $\psi_k$ corresponding to the rays.	
The instance field is supervised using a margin-based contrastive objective. Specifically, two rays, $r_i$ and $r_j$, are cast from the pixels in the masks $M_i$ and $M_j$, with physical scales $S_i$ and $S_j$.	
The instance features $\psi_i$ and $\psi_j$ are obtained by volumetrically rendering the embeddings along their respective rays $r_i$ and $r_j$.	
The concrete contrastive loss $L_{in}$ is formulated as: 	
\begin{equation}\label{eq5}
	L_{in} = \left\{ 
	\begin{aligned}
		& \lVert \psi_i - \psi_j \rVert  && \text{if } M_i = M_j, \\
		& \text{ReLU}(\lambda_{in} - \lVert \psi_i - \psi_j \rVert) && \text{if } M_i \neq M_j, \\
	\end{aligned}\right.
\end{equation}
where $\lambda_{in}$ represents the bound constant as a hyperparameter.	
It is important to note that this instance loss is applied to rays sampled from the same viewpoint.	
This supervision allows the instance field to differentiate instances with identical language features and further enhance localization precision.	

\noindent \textbf{Analysis} 
Opacity and color, as essential visual properties extracted from scene reconstruction, provide rich geometric and appearance information that is effectively integrated into the hierarchical feature field construction. 
This enhances the quality of language embeddings, improving the accuracy of 3D visual grounding and spatial reasoning.
Consequently, SpatialReasoner can effectively learn to represent high-quality language and instance features with complex characteristics.	
The effectiveness of the visual properties will be demonstrated in the experiments (Sec.~\ref{p4}).	

SpatialReasoner is a versatile framework that constructs hierarchical feature fields using diverse neural representations, including Neural Radiance Fields (NeRF) for photorealistic scene modeling, Instant Neural Graphics Primitives (Instant-NGP) for fast training and rendering, and 3D Gaussian Splatting (3DGS) for efficient, scalable scene representation. This hierarchical construction enables SpatialReasoner to model 3D space through language-instance interactions, facilitating spatial reasoning and precise instance localization within complex 3D scenes.

\subsection{Spatial Reasoning and Grounding}
Through Spatial Relation Decomposition (``\textbf{think carefully}'') and Hierarchical Feature Field Construction (``\textbf{look carefully}''), SpatialReasoner has acquired spatial relationships embedded in complex text queries and a comprehensive scene representation, including geometry and appearance, language features, and instance features.	
Subsequently, SpatialReasoner will execute spatial reasoning and grounding to localize the target instance step-by-step.

\noindent \textbf{Relevancy Map Activation.} SpatialReasoner utilizes reasoned instructions from the finetuned LLM to hierarchically query the hierarchical feature field.	
Based on the reasoned target and anchor instructions, we first query the language field using spatial coordinates, physical scales, and visual properties.	
Consider that the physical scales $\{S_1, S_2, ..., S_n\}$ of objects in 3D space and the CLIP embeddings $\{\phi_1, \phi_2, ..., \phi_n\}$ are established during the supervision generation process. Additionally, the targets and anchors are bound to appear in the training images.	 
To this end, the CLIP embeddings are used to compute the similarity with the CLIP textual features of the target and anchor instructions individually.	
The physical scales corresponding to the candidate with the highest similarity are used to query the language field and generate the language embedding maps.	
Thus, SpatialReasoner activates the relevance map through the similarity computation between the language embedding maps and the CLIP textual features of the target and anchor instructions.	

\noindent \textbf{Candidate Generation.} 
Based on the relevance map, SpatialReasoner determines the relevant region candidates $     \{C_1, C_2, ..., C_n\}$ related to the language feature of target and anchor instructions, depending on the maximum relevance scores.	
To differentiate instances with similar language features and further refine localization precision, SpatialReasoner deprojects the relevant regions into 3D space for scale calculation and queries the hierarchical instance field for instance feature acquisition.	
SpatialReasoner deprojects the pixel with the maximum relevance from each relevant region to the 3D point and obtains the candidate instance feature.	

\noindent \textbf{Instance Graph Construction.} Candidates determined based solely on language are often incomplete.	
We need to merge the candidates based on the instance feature.	
SpatialReasoner constructs the instance graph $G(V,E)$ to prune and merge the candidate set $\{C_1, C_2, ..., C_n\}$.	
The node set $V = \{v_1, v_2, ..., v_n\}$ of the graph includes the features of the candidate objects.	
Next, we compute the affine differences $A$ within the node set by measuring the Euclidean distances between node pairs:	
\begin{equation}\label{eq_44}
	\begin{aligned}
		&A = \left\lVert V {\otimes} \mathbf{1}_n - (V {\otimes} \mathbf{1}_n)^T \right\rVert_2
	\end{aligned}
\end{equation}
where $\mathbf{1}_n$ is a vector of all ones with size $n \times 1$ and $\otimes$ denotes the tensor product.	
The edge set $E = \{e_{ij}\}$ is defined based on the differences $A$.	
Thus, $G(V, E)$ can be used to determine its connected components.	
The set of all connected components of the graph is denoted as $\{N_1, N_2, \ldots, N_k\}$.	
Each connected component $N_i$ merges the candidates into the final complete candidate $\hat{C}_i$ using a bitwise OR operation:	
\begin{equation}\label{eq_6}
	\begin{aligned}
		& \hat{C}_i = \vee_{j \in N_i} C_j
	\end{aligned}
\end{equation}

This step combines the candidates associated with each connected component into the final set of candidates $\{\hat{C}_1, \hat{C}_2, \ldots, \hat{C}_k\}$.	
To localize specific instances from the candidates based on the reasoned spatial relation instructions, SpatialReasoner first determines the target and anchor candidates as described above.	
We deproject the target and anchor candidates into 3D space and warp them with bounding boxes.	
SpatialReasoner considers four spatial relations (Horizontal Proximity, Vertical Proximity, Support, Allocentric) to localize the specific instance.	
Generally, the language and instance fields are queried in a hierarchical manner based on the target and anchor instructions, and the specific object is determined accordingly with the spatial relation instructions.	
This enables SpatialReasoner to perform spatial reasoning and grounding within both 3D scenes and language queries, resulting in the accurate localization of specific instances mentioned in the language queries.

\section{Experiments}
\label{p4}

\subsection{Experimental Settings}
\label{4.1}
\noindent \textbf{Datasets.}
For quantitative and qualitative validation, we conduct experiments on the LERF~\cite{kerr2023lerf} dataset, the Replica~\cite{straub2019replica} dataset, and our developed Re3D dataset.	
Additionally, we fine-tune and evaluate an LLM on the Sr3D and Sr3D++ datasets~\cite{achlioptas2020referit3d}.	

\begin{itemize}
    \item \textbf{LERF dataset.} The LERF dataset comprises real-world and posed long-tail scenes captured via the Polycam iPhone app, which employs on-board SLAM for camera pose estimation and feature extraction. SpatialReasoner utilizes an extended version of LERF with annotated ground truth, further augmented to evaluate 3D spatial reasoning and open-vocabulary visual grounding. To assess localization performance under spatial language queries, we enhance LERF by rendering novel views and generating ground truth labels using the Segment Anything Model (SAM).
    \item \textbf{Replica dataset.}
    The Replica dataset consists of high-quality reconstructions of various indoor spaces.	
    Each reconstruction features clean, dense geometry, high-resolution, high-dynamic-range textures, glass and mirror surface information, planar segmentation, and semantic class and instance segmentation.	
    Each scene is particularly well-suited for evaluating open-set 3D scene understanding.	
    \item \textbf{Re3D dataset.} To further evaluate the capability of the proposed SpatialReasoner in localizing specific instances by understanding complex sentence queries containing spatial relations, we developed the Re3D dataset, which mainly involves various objects with the same semantics but differing spatial conditions.	
    Similarly, the Re3D dataset is also captured using the Polycam, employing on-board SLAM.	    
\end{itemize}

\begin{figure*}
	\centering
    \includegraphics[width=0.9\textwidth]{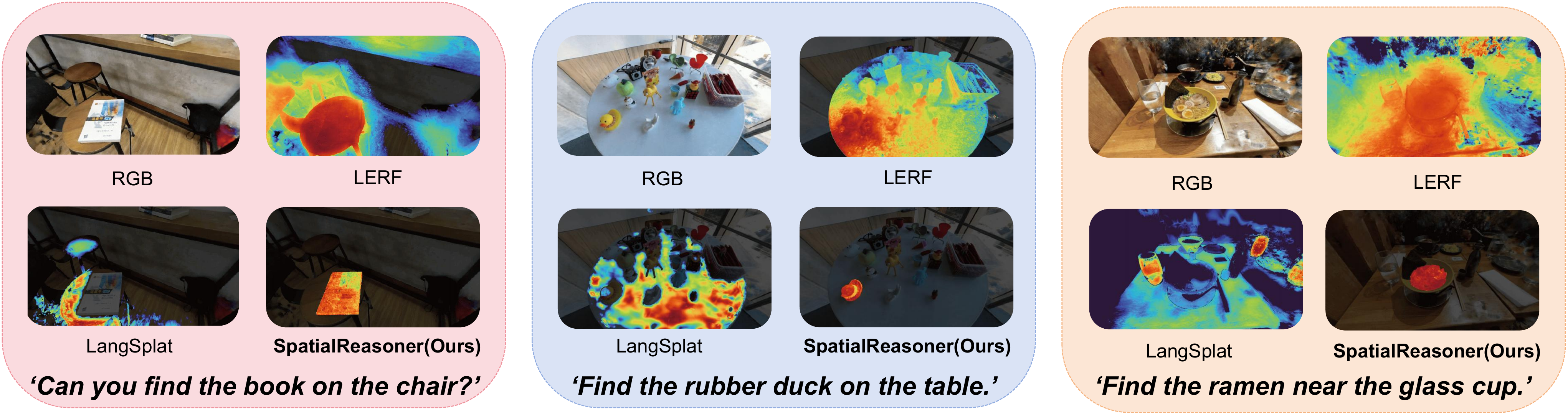}
	\caption{Qualitative comparisons of spatial reasoning capability. Results demonstrate that our SpatialReasoner achieves spatial reasoning and localizing the target instance based on the spatial relation.\label{qual_1}}
\end{figure*}

\noindent \textbf{Implementation Details.} The Segment Anything Model~\cite{kirillov2023segment} is used to generate the segmentation masks from the multi-view training images.	
The TinyLlama~\cite{zhang2024tinyllama} model is used to comprehend the spatial relation within the language query to ``think carefully''.	
The Adam optimizer with a weight decay of 1e-9 is used for the proposal networks and fields.	
Training involves an exponential learning rate scheduler, transitioning from 1e-2 to 1e-3 during the first 5000 steps.	
All experiments are conducted on NVIDIA H100 GPUs.	

\noindent \textbf{Comparison Methods.} Given the scarcity of existing methods for 3D spatial reasoning and grounding, we selected several representative approaches for comparison.	Specifically, we compared our method with the following state-of-the-art methods: open-vocabulary 3D visual grounding methods LERF and LangSplat, and the open-vocabulary 2D semantic segmentation methods ODISE~\cite{xu2023open}, OV-Seg~\cite{liang2023open}, and LSeg~\cite{li2022language}.	To validate the generality of our SpatialReasoner, we further integrate the proposed SpatialReasoner into various neural representations: Neural Radiance Fields (NeRF), Instant-NGP (NGP), and 3D Gaussian Splatting (3DGS), which are denoted as \textbf{SpatialReasoner(NeRF)}, \textbf{SpatialReasoner(NGP)}, and \textbf{SpatialReasoner(3DGS)}, respectively.	

\noindent \textbf{Metrics.} The experiments mainly utilize localization accuracy and mean intersection over union (mIoU) for evaluation.
Localization accuracy considers a label a success if the highest relevance pixel falls inside the annotated box.	
The mIoU calculates the intersection ratio between the rendered relevance map and the annotated ground truth to measure the localization accuracy of the region.	

\begin{table}[t]
	\renewcommand\tabcolsep{1.5pt}
	\centering
\caption{\textbf{Localization accuracy scores (\%) on LERF dataset for spatial reasoning.} The first three methods target 2D domain, whereas the remaining methods, including our SpatialReasoner, focus on 3D domain.}
	\label{table:1}
    \vspace{-0.10in}

	\begin{tabular}{lccccc}
		\toprule
		\textbf{Method} & \textit{ramen}   & \textit{figurines} & \textit{teatime}  & \textit{kitchen}  & \textbf{overall}   \\
		\midrule
		LSeg~\cite{li2022language} & 11.4 & 12.2 & 10.4 & 10.6 & 11.2 \\
		ODISE~\cite{xu2023open} & 12.1 & 21.8 & 23.6 & 21.3 & 19.7 \\
		OV-Seg~\cite{liang2023open} & 21.5 & 22.1 & 23.8 & 20.4 & 21.9\\
		\midrule

		LERF~\cite{kerr2023lerf} & 32.4 & 44.3 & 40.8 & 33.1 & 37.6\\
		LangSplat~\cite{qin2023langsplat} & 45.4 & 31.2 & 58.7 & 48.8 & 46.1 \\ 
        \midrule
		\textbf{SpatialReasoner(NeRF)} & \cellcolor{softyellow}\textbf{89.7} & \cellcolor{softyellow}\textbf{93.2} & \cellcolor{softyellow}\textbf{64.6} & \cellcolor{softyellow}\textbf{86.2} & \cellcolor{softyellow}\textbf{83.4} \\ 	
		\textbf{SpatialReasoner(NGP)} & \cellcolor{softorange}\textbf{92.8} & \cellcolor{softorange}\textbf{97.7} & \cellcolor{softorange}\textbf{72.4} & \cellcolor{softorange}\textbf{88.4} & \cellcolor{softorange}\textbf{87.8} \\
		\textbf{SpatialReasoner(3DGS)} & \cellcolor{softred}\textbf{94.1} & \cellcolor{softred}\textbf{98.3} & \cellcolor{softred}\textbf{83.2} & \cellcolor{softred}\textbf{91.5} & \cellcolor{softred}\textbf{91.7} \\
		\bottomrule
	\end{tabular}
    \vspace{-0.15in}
\end{table}

\begin{table}[t]
	\renewcommand\tabcolsep{1.5pt}
	\centering
\caption{\textbf{mIoU scores (\%) on LERF dataset for spatial reasoning.} The first three methods target 2D, whereas the remaining methods, including our SpatialReasoner, focus on 3D.}
	\label{table:2}
    \vspace{-0.10in}
	\begin{tabular}{lccccc}
		\toprule
		\textbf{Method} & \textit{ramen}   & \textit{figurines} & \textit{teatime}  & \textit{kitchen}  & \textbf{overall}   \\
		\midrule
		LSeg~\cite{li2022language} & 14.4 & 16.0 & 19.2 & 24.5 & 18.5 \\
		ODISE~\cite{xu2023open} & 22.4 & 14.1 & 22.5 & 28.4 & 21.8 \\
		OV-Seg~\cite{liang2023open} & 33.2 & 18.9 & 21.4 & 26.4 & 24.9\\
		\midrule

		LERF~\cite{kerr2023lerf} & 53.4 & 46.2 & 48.5 & 46.8& 48.7 \\
		LangSplat~\cite{qin2023langsplat} & 58.5 & 59.2 & 54.4 & 63.2 &  58.8 \\ 
        \midrule
		\textbf{SpatialReasoner(NeRF)} & \cellcolor{softyellow}\textbf{87.2} & \cellcolor{softyellow}\textbf{89.6} & \cellcolor{softyellow}\textbf{75.5} & \cellcolor{softyellow}\textbf{88.7} & \cellcolor{softyellow}\textbf{85.3} \\ 	
		\textbf{SpatialReasoner(NGP)} & \cellcolor{softorange}\textbf{93.2} & \cellcolor{softorange}\textbf{95.6} & \cellcolor{softorange}\textbf{84.2} & \cellcolor{softorange}\textbf{90.6} & \cellcolor{softorange}\textbf{90.9} \\
		\textbf{SpatialReasoner(3DGS)} & \cellcolor{softred}\textbf{93.6} & \cellcolor{softred}\textbf{97.4} & \cellcolor{softred}\textbf{86.5} & \cellcolor{softred}\textbf{93.7} & \cellcolor{softred}\textbf{92.8} \\
		\bottomrule
	\end{tabular}
    \vspace{-0.20in}
\end{table}

\begin{table}
	\renewcommand\tabcolsep{6pt}
	\centering
    \caption{\textbf{mIoU scores (\%) on Replica dataset for spatial reasoning.} The first three methods target 2D, whereas the remaining methods, including our SpatialReasoner, focus on 3D. \label{table:3}}
    \vspace{-0.10in}
	\begin{tabular}{lcccc}
		\toprule
		\textbf{Methods} & \textit{head} & \textit{common} & \textit{tail} & \textbf{overall} \\
		\midrule
		LSeg~\cite{li2022language} & 7.1  &6.6 & 1.4& 5.1\\
        ODISE~\cite{xu2023open} & 10.8  &8.2 & 1.2 & 6.7\\
        OV-Seg~\cite{liang2023open} &13.4  &7.7 &1.4 & 7.5\\
        \midrule
		LERF~\cite{kerr2023lerf} & 19.2 & 10.1 & 2.3 & 10.5\\
        LangSplat~\cite{qin2023langsplat} & 22.3  & 15.2 & 4.3 & 13.9\\
         
       \midrule
	\textbf{SpatialReasoner(NeRF)} & \cellcolor{softyellow}{\textbf{22.8}} &\cellcolor{softyellow}{\textbf{16.7}} & \cellcolor{softyellow}\textbf{6.2} &\cellcolor{softyellow}\textbf{15.2}\\
    \textbf{SpatialReasoner(NGP)} & \cellcolor{softorange}{\textbf{26.5}} &\cellcolor{softorange}{\textbf{28.3}} & \cellcolor{softorange}\textbf{10.2}&\cellcolor{softorange}\textbf{21.6} \\
    \textbf{SpatialReasoner(3DGS)} & \cellcolor{softred}\textbf{31.7} &\cellcolor{softred}\textbf{36.6} & \cellcolor{softred}\textbf{18.4} & \cellcolor{softred}\textbf{28.9}\\
		\bottomrule
	\end{tabular}
    \vspace{-0.15in}
\end{table}

\begin{table}
	\renewcommand\tabcolsep{6pt}
	\centering
    \caption{\textbf{Quantitative results on Re3D dataset for spatial reasoning.} Acc denotes the localization accuracy, mIoU denotes the mean IoU score, and Speed denotes the activation time for a single view. \label{table:4}}
	\vspace{-0.10in}
    \begin{tabular}{lccc}
		\toprule
		\textbf{Methods} & \textbf{Acc(\%)} & \textbf{mIoU(\%)} & \textbf{Speed(s)}\\
		\midrule
		LSeg~\cite{li2022language} & 24.1  &18.6 & 1.43 \\
        ODISE~\cite{xu2023open} & 32.4  &33.6 & 1.22 \\
        OV-Seg~\cite{liang2023open} &38.4  &34.8 &1.23 \\
        \midrule
		LERF~\cite{kerr2023lerf} & 71.4 & 44.6 & 0.93\\
        LangSplat~\cite{qin2023langsplat} & 85.6  & 51.2 & \cellcolor{softred}0.04 \\
         
       \midrule
		\textbf{SpatialReasoner(NeRF)} & \cellcolor{softyellow}{\textbf{85.8}} &\cellcolor{softyellow}{\textbf{52.4}} & \textbf{1.43} \\
       \textbf{SpatialReasoner(NGP)} & \cellcolor{softorange}{\textbf{91.4}} &\cellcolor{softorange}{\textbf{62.6}} & \cellcolor{softyellow}\textbf{0.44} \\
    \textbf{SpatialReasoner(3DGS)} & \cellcolor{softred}\textbf{92.6} &\cellcolor{softred}\textbf{64.8} & \cellcolor{softorange}\textbf{0.08} \\
		\bottomrule
	\end{tabular}
    \vspace{-0.05in}
\end{table}

\begin{table}
	\renewcommand\tabcolsep{3pt}
	\centering
    \caption{\textbf{Localization accuracy scores on extended LERF dataset (LERF-ex), Re3D dataset, and Replica dataset for spatial reasoning.} The first three methods target 2D domain, whereas the remaining methods, including our SpatialReasoner, focus on 3D domain.
\label{table:5}}
    \vspace{-0.10in}
	\begin{tabular}{lcccc}
		\toprule
		\textbf{Methods} & \textbf{LERF-ex} & \textbf{Re3D} & \textbf{Replica} & \textbf{overall} \\
		\midrule
		LSeg~\cite{li2022language} &21.6  &34.8 & 17.2 & 24.5\\
        ODISE~\cite{xu2023open} & 30.2  &43.8 & 28.6 & 34.2\\
        OV-Seg~\cite{liang2023open} &35.4  &52.6 &30.2 &39.4\\
        \midrule
		LERF~\cite{kerr2023lerf} & 74.8 & 45.4 & 32.2& 50.8\\
        LangSplat~\cite{qin2023langsplat} & 85.4  & 56.3 & 42.6 & 61.4\\
         
       \midrule
	\textbf{SpatialReasoner(NeRF)} & \cellcolor{softyellow}{\textbf{91.8}} &\cellcolor{softyellow}{\textbf{67.6}} & \cellcolor{softyellow}\textbf{51.2} &\cellcolor{softyellow}\textbf{70.2}\\
    \textbf{SpatialReasoner(NGP)} & \cellcolor{softorange}{\textbf{93.2}} &\cellcolor{softorange}{\textbf{72.4}} & \cellcolor{softorange}\textbf{59.4} & \cellcolor{softorange}\textbf{75.0}\\
    \textbf{SpatialReasoner(3DGS)} & \cellcolor{softred}\textbf{94.5} &\cellcolor{softred}\textbf{76.8} & \cellcolor{softred}\textbf{68.4} &\cellcolor{softred}\textbf{79.9}\\
		\bottomrule
	\end{tabular}

\end{table}

\begin{figure*}
	\centering
    \includegraphics[width=0.9\textwidth]{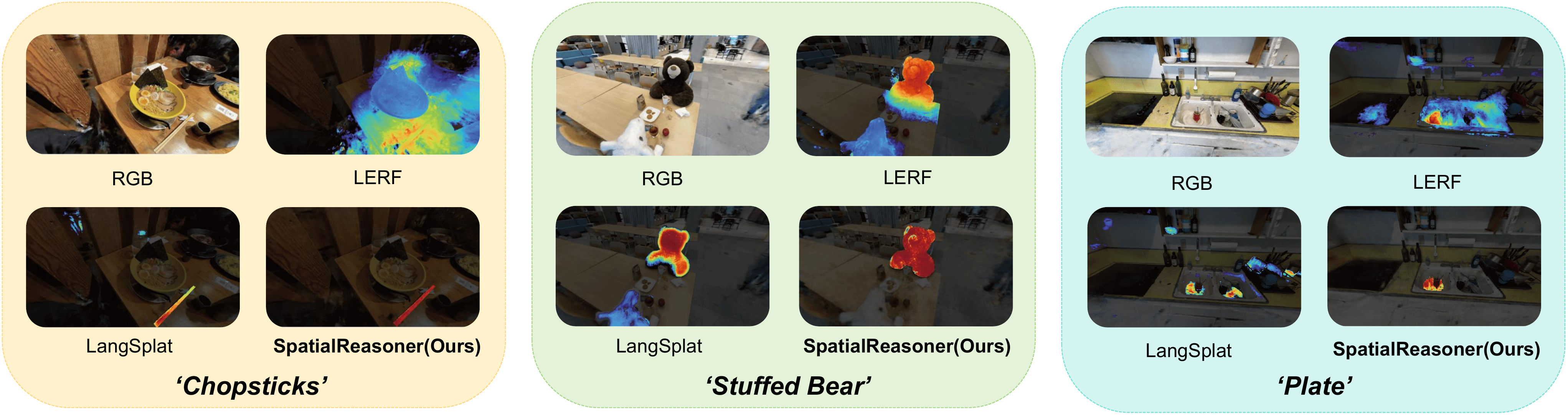}
	\caption{Qualitative comparisons of 3D visual grounding capability. Results demonstrate that our SpatialReasoner achieves the superior accuracy in open-vocabulary 3D localization compared to other state-of-the-art methods.\label{qual_2}}
    
\end{figure*}

\subsection{Results of Spatial Reasoning}
\noindent \textbf{Quantitative Results.} 
We compare our method with the comparison methods on the extended LERF, Replica, and Re3D datasets.	
As illustrated in Table~\ref{table:1} and Table~\ref{table:2}, our method achieves an overall localization accuracy of 83.4\%, 87.8\%, and 91.7\%, and an mIoU of 85.3\%, 90.9\%, and 92.8\% by integrating SpatialReasoner into NeRF, Instant-NGP, and 3DGS, significantly outperforming the comparison methods.	
The Replica dataset is categorized into head, common, and tail based on the number of annotated points.	
As illustrated in Table~\ref{table:3}, our method outperforms the comparison methods across all categories.	
To further evaluate the capability for spatial reasoning, we compare our method with other methods on the Re3D dataset.	
Our method significantly improves the performance without significantly increasing the inference cost, as shown in Table~\ref{table:4}.

\noindent \textbf{Qualitative Results.} Figure~\ref{qual_1} illustrates the spatial reasoning results.	
SpatialReasoner demonstrates a robust spatial reasoning capability, enabling it to localize instances based on the spatial relations in language queries.	
It provides accurate 3D localozation even in complex real-world scenes.	
For instance, given the query ``Can you find the book on the chair?'' SpatialReasoner accurately localizes the book on the chair, whereas LERF and LangSplat fail to handle this query involving spatial reasoning.	

\subsection{Results of Grounding}
\textbf{Quantitative Results.}
SpatialReasoner can be integrated into different neural representations, 
outperforming existing methods on 3D visual grounding. As demonstrated in Table.\ref{table:5}, our method achieves an overall localization accuracy of 70.2\%, 75.0\%, and 79.9\% by integrating SpatialReasoner into NeRF, Instant-NGP, and 3DGS, significantly outperforming the comparison methods.

\noindent \textbf{Qualitative Results.} We show the high quality of open-vocabulary 3D visual grounding in SpatialReasoner compared to other methods in challenging and realistic scenes. Figure~\ref{qual_2} illustrates the visual results of this comparison. We note that the activation areas generated by other methods are often more scattered, while our approach yields more accurate regions. 

\begin{table}[t]
	\vspace{-6pt}
	\renewcommand\tabcolsep{4pt} 
	\centering
    \caption{Ablation studies of key components on LERF dataset.
    We report the mIoU score (\%) for performance evaluation and inference time (s) for computation cost evaluation.
    }\label{table:ablation}
	\begin{tabular}{ccc|cc}
		\toprule
		\multicolumn{3}{c|}{\textbf{Component}} & \multicolumn{2}{c}{\textbf{Performance (mIoU / Inference Time)}} \\
		\midrule
		\textbf{VP} & \textbf{HIF} & \textbf{IG} & \textbf{3D Visual Grounding} & \textbf{Spatial Reasoning} \\
		\midrule
		&  &  & 78.2/0.038 & \XSolidBrush \\
		\CheckmarkBold &  &  & 82.4/0.042 & \XSolidBrush  \\
		\CheckmarkBold & \CheckmarkBold &  & 88.3/0.051 & 86.9/0.072 \\
		\CheckmarkBold & \CheckmarkBold & \CheckmarkBold & {94.5/0.063} & {91.7/0.091} \\
		\bottomrule
	\end{tabular}	
\end{table}

\begin{figure}
	\centering	\includegraphics[width=1.0\linewidth]{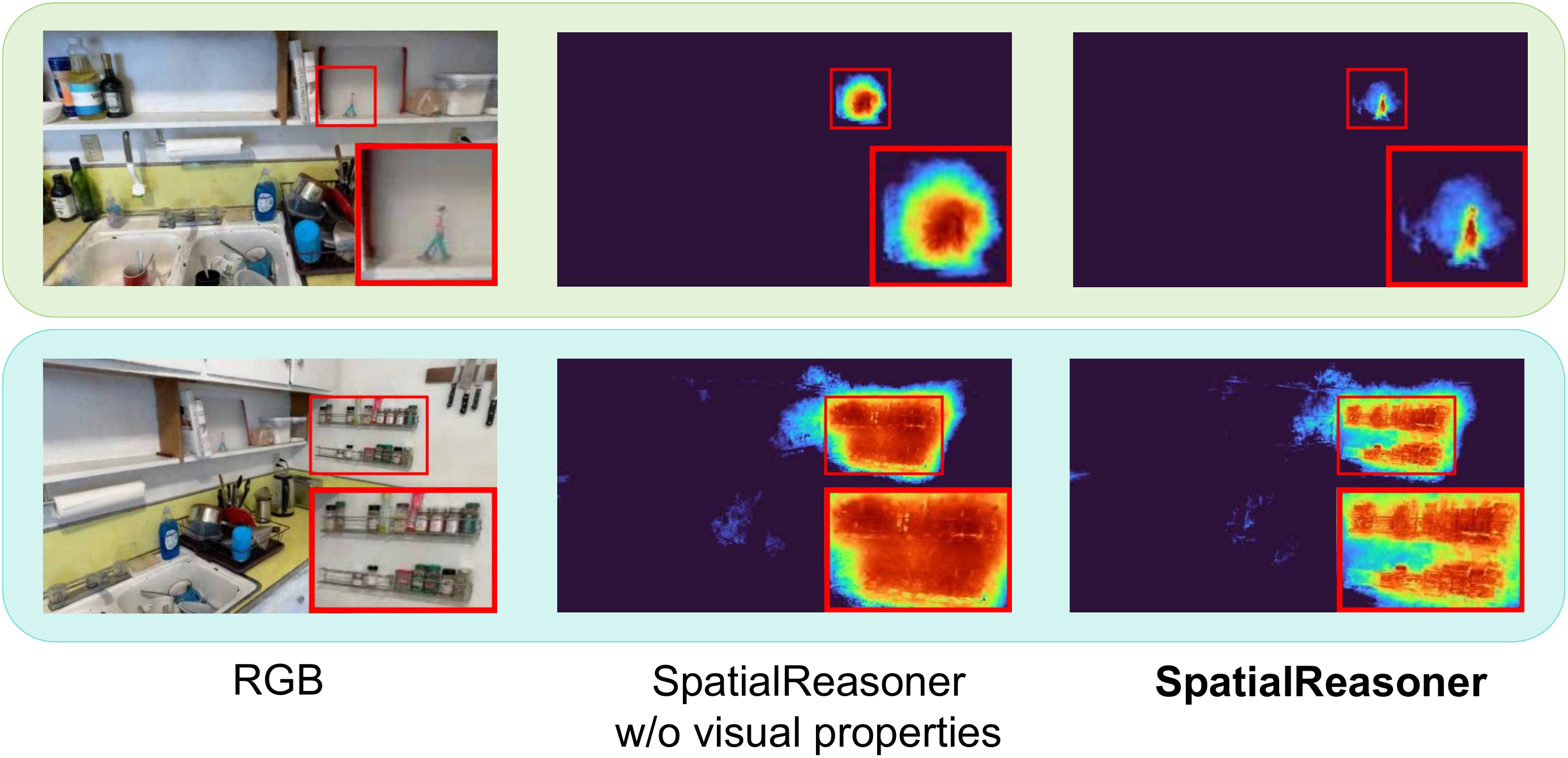}
	\caption{Ablation Study on visual properties with qualitative degradation examples.  \label{ab1}}
\end{figure}

\begin{figure}
	\centering	\includegraphics[width=1.0\linewidth]{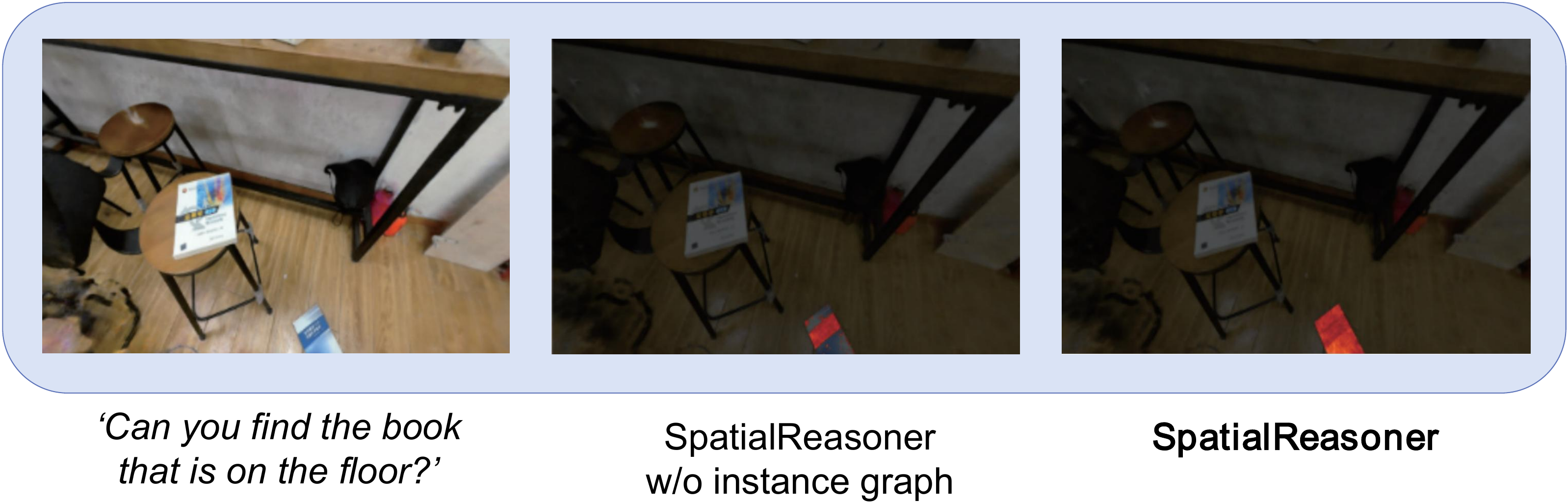}
	\caption{Ablation Study on instance graph construction with qualitative degradation examples.
    \label{ab2}}
    \vspace{-0.15in}
\end{figure}
\subsection{Ablation Study}
Ablations are conducted on the LERF dataset with SpatialReasoner(3DGS) in Table~\ref{table:ablation}.
Without the proposed components, we only utilize the hierarchical language field to enable open-vocabulary 3D visual grounding. It achieves a mIoU score of 78.2\% and renders each view in 0.038 seconds.
Through the introduction of visual properties (\textbf{VP}) from scene reconstruction, SpatialReasoner further enhance the performance of 3D visual grounding by 4.2\%. 
Furthermore, SpatialReasoner introduces a hierarchical instance field (\textbf{HIF}) for spatial reasoning, enabling 3D localization with spatial relations.
By incorporating the instance graph construction (\textbf{IG}), SpatialReasoner further increases the performance of grounding by 6.2\% and spatial reasoning by 4.6\%.
We also provide qualitative comparison results in Figure~\ref{ab1} and Figure~\ref{ab2}.

\section{Conclusion}
\label{p6}
We propose a novel neural representation-based framework with LLM-driven spatial reasoning, SpatialReasoner, for open-vocabulary 3D visual grounding.	
SpatialReasoner leverages an LLM for spatial relation decomposition, along with visual property-enhanced hierarchical feature fields for spatial reasoning, allowing it to ``think carefully'' and ``look carefully,'' enabling accurate, step-by-step localization of the target instance.	
Extensive experiments demonstrate that our method can be seamlessly integrated into various 3D neural representations, surpassing baseline models in 3D visual grounding and empowering their spatial reasoning capabilities.	

\section{Acknowledgments}
This work was supported in part by NSFC Project (62176061), Science and Technology Commission of Shanghai Municipality (No.24511103100), Doubao Fund, and Shanghai Technology Development and Entrepreneurship Platform for Neuromorphic and AI SoC. The authors gratefully acknowledge the support and resources provided by these organizations

\bibliographystyle{ACM-Reference-Format}
\bibliography{sample-base}
\end{document}